# From Research Frontier to Laboratory Bench: Design of a Four-Tier Experimental Teaching System for Multimodal Medical Image Intelligent Diagnosis

Dongjing Shan[1], Yamei Luo[1], Jin Li[1], Yong Luo[2]

[1] *School of Medical Information and Engineering, Southwest Medical University, Luzhou 646000, Sichuan, China*

[2] *National Engineering Research Center for Multimedia Software, School of Computer Science, Wuhan University, Wuhan 430079, Hubei, China*

*Corresponding authors: shandongjing@swmu.edu.cn, yluo180@gmail.com*

**Abstract.** Undergraduate programmes in intelligent medical engineering are expanding, yet laboratory curricula lag behind the multimodal, long-tailed, and distributionally shifting realities of clinical AI. This design paper presents an advanced experimental teaching system that translates an ongoing multimodal deep learning research project on endometrial carcinoma into a structured undergraduate lab sequence. We identify three educational gaps (modality, authenticity, and deployment) and derive four pedagogical principles from constructive alignment, experiential learning, the research teaching nexus, and the CDIO framework. The curriculum comprises four progressive tiers plus an engineering layer, with 32 laboratory units over 64 contact hours, delivered via a custom virtual clinical workstation using de-identified multi-institutional data. Each tier maps to a specific technical bottleneck, prerequisite coursework, and criterion-referenced deliverables. Data governance, safety, and assessment protocols are specified. Learning outcome data will be collected across two implementation cycles.

**Keywords.** intelligent medical engineering; experimental teaching; multimodal deep learning; research–teaching nexus; curriculum design; medical AI education; constructive alignment.

## 1. Introduction

Medical AI has advanced rapidly, evolving from proof-of-concept classifiers to regulated systems deployed in clinical practice [1], [2]. Yet undergraduate curricula for this field have lagged substantially. Despite calls for holistic AI-centred restructuring [3], systematic reviews indicate that AI instruction remains disconnected from clinical reasoning [4], [5]. This gap is most pronounced in laboratory teaching, which resists rapid modernisation due to dependence on datasets, hardware, toolchains, and safety protocols.

At Southwest Medical University, the intelligent medical engineering programme, launched in 2022, confronted this challenge when designing its laboratory sequence. We identified a fundamental misalignment. Advanced techniques prevalent in current research, such as multimodal fusion, graph-based representation learning, continual learning, and robustness to distribution shift, find no natural place in conventional single-modality image-processing labs. Traditional exercises rely on curated, balanced datasets and fixed protocols with predetermined outcomes, failing to prepare students for real-world deployment where models may degrade on unseen institutional data or must operate on resource-constrained devices such as ultrasound carts.

In response, we developed an advanced experimental teaching system for multimodal medical image intelligent diagnosis, in collaboration with the National Engineering Research Center for Multimedia Software at Wuhan University. Grounded in active research on multimodal deep learning for endometrial carcinoma [21]-[27], together with related work on medical image analysis and graph-based classification [28]-[31], the system comprises four progressive tiers (representation learning, fusion and reasoning, knowledge evolution, and robustness under long-tailed and out-of-distribution conditions), encompassing 32 laboratory units across 64 contact hours. Infrastructure includes a virtual clinical diagnosis workstation, criterion-referenced assessment, and a rigorous data governance framework enabling safe undergraduate access to de-identified multi-institutional clinical data.

Our contribution is threefold. First, we decompose the gap between current clinical practice and laboratory education into three structural deficits: modality, authenticity, and deployment, each warranting distinct pedagogical interventions. Second, we anchor every laboratory tier to a recognised unsolved technical bottleneck in the literature, ensuring the curriculum is driven by enduring problems rather than transient tools. Third, we specify institutional infrastructure, including technological platforms, data governance protocols, staffing, and assessment methods, in sufficient detail to support adoption or adaptation across academic settings..

## 2. Three Structural Gaps in Medical AI Laboratory Teaching

**The modality gap**. Clinical decisions about a tumour draw on ultrasound, contrast-enhanced ultrasound, CT or MRI, molecular pathology, and free-text clinical notes; the diagnostic value of the ensemble exceeds any single component [6], [7]. Laboratory teaching is overwhelmingly single-modality, so students lack practice at fusion. The characteristic failure modes of multimodal learning are invisible to them. Modality laziness, the tendency of a jointly trained network to converge onto whichever modality is easiest to fit and leave the others under-optimised [9], cannot be observed or corrected by a student who has never trained a joint model. Similarly, semantic misalignment between heterogeneous encoders and noise propagation from a low-quality modality into a fused representation remain unseen.

**The authenticity gap.** Teaching datasets are curated to be tractable: balanced, complete, single-source and accompanied by settled evaluation protocols. Clinical data are none of these. Rare histological subtypes occupy the tail of a long-tailed distribution [13]; scanner, protocol and population differences produce distribution shift between institutions; modalities are missing for many patients; and the correct evaluation protocol is itself a matter of judgement. A student who has only worked with curated data learns that improving a metric is the object of the exercise. The habits clinical work requires, asking whether a gain survives a change of institution, whether a model can decline to answer on out-of-distribution input [14], and whether a threshold is defensible, are never elicited because curated datasets never punish their absence.

**The deployment gap**. Medical AI is most needed where computation is least available, such as county hospitals, community clinics, and portable ultrasound and MRI units. Laboratory teaching almost universally ends at the validation metric. Model compression, quantisation, latency budgets and cross-platform adaptation are treated as industrial afterthoughts rather than design constraints shaping architecture choice. A related omission is interpretability: saliency and attention-attribution methods [11], [12] are taught as visualisation utilities but rarely as instruments students must use to justify a decision to a clinician who will not accept the output otherwise.

These three gaps call for different remedies. The modality gap is closed by content: the curriculum must contain fusion experiments that cannot be completed with a single modality. The authenticity gap is closed by data and assessment: students must work with de-identified multi-institutional data, and assessment criteria must reward robustness rather than peak accuracy. The deployment gap is closed by infrastructure: an edge target must exist in the laboratory, and a latency budget must form part of the pass criterion. A curriculum addressing only one gap leaves the other two intact..

## 3. Design Principles and System Architecture

### *3.1 Design principles*

Four principles governed the design, each drawn from an established position in the education literature and each translated into a testable constraint on the artefact.

**Constructive alignment.** Biggs argues that intended learning outcomes, learning activities and assessment tasks must be mutually consistent, and that misalignment, not student ability, explains much of the observed shortfall in higher education [15]. Applied here, the constraint is that no laboratory unit enters the system unless it is paired with an explicit outcome statement and an assessment task that cannot be passed without achieving the outcome. This is why our thresholds are criterion-referenced and stated numerically in advance (Table 3) rather than left to holistic judgement.

**Experiential progression.** Kolb's cycle of concrete experience, reflective observation, abstract conceptualisation and active experimentation [16], and the evidence that active engagement outperforms passive exposure in science and engineering [19], imply that a laboratory sequence should return repeatedly to the same construct at increasing levels of abstraction rather than treating each topic once. Our tiers are therefore cumulative: the graph representation built in Tier I is the object that Tier II fuses, that Tier III must preserve across tasks, and that Tier IV must keep stable under shift.

**The research–teaching nexus.** Healey distinguishes curricula in which students hear about research from those in which they participate in it [20], and only the latter reliably develops research disposition. We therefore transfer not only results from the parent research programme but also its open problems. Where the research has a settled answer, students reproduce it; where it does not, students are asked to propose and test an improvement, and a negative result is an acceptable outcome provided the experiment was sound.

**Conceive–design–implement–operate.** The CDIO framework [18] insists that engineering education be organised around the full product lifecycle. The practical consequence is the existence of Tier V, the engineering layer: a student who has trained an accurate model has completed only half of the exercise, and the remaining half (compression, quantisation, latency measurement, interpretability reporting and multi-centre verification) is separately weighted in the mark, as the translation stage of Table 3, rather than treated as an optional extension.

### *3.2 Four progressive tiers and an engineering layer*

The spine of the system is a sequence of four tiers, each anchored to a bottleneck that is genuinely open in the multimodal medical imaging literature, followed by an engineering layer that carries the artefact to a deployable state. Table 1 gives the mapping from bottleneck to tier to prerequisite coursework. Tier content follows the corresponding methodological literature: graph attention for fusion [8], gradient modulation for balanced training [9], and regularisation against catastrophic forgetting [10].

Table 1. Mapping from open technical bottlenecks to laboratory tiers, prerequisite coursework and terminal deliverables.

| Tier | Bottleneck addressed | Core laboratory content | Prerequisite courses | Terminal deliverable |
|---|---|---|---|---|
| I. Representation learning | Semantic gap and feature degeneration across heterogeneous modalities | Graph-structured unified encoding of ultrasound, MRI and pathology text; dynamic topology bootstrapping; low-rank and sparse optimisation | Machine Learning; Medical Image Processing | Unified multimodal encoding with a reproducible topology |
| II. Fusion and reasoning | Dynamic coupling mismatch, missing modalities, cross-modal noise propagation | Graph attention networks; gated recurrent graph networks; sparsity-aware attention; missing-modality completion | Machine Learning; Biomedical Big Data | Fusion model with an ablation study over modality subsets |
| III. Knowledge evolution | Modality laziness and catastrophic forgetting under streaming data | Orthogonal subspace projection and prompt learning; rule-based cross-modal transfer; gradient modulation for balanced training | Deep Learning; Pattern Recognition | Incremental-learning report with forgetting-rate curves |
| IV. Robustness | Label scarcity, long-tailed subtypes, distribution shift, out-of-distribution inputs | Contrastive pretraining; re-weighted adversarial sample generation; manifold adversarial training; OOD detection and abstention | Deep Learning; Medical Statistics | Robustness dossier covering tail-class and OOD evaluation |

| Tier | Bottleneck addressed | Core laboratory content | Prerequisite courses | Terminal deliverable |
|---|---|---|---|---|
| V. Engineering layer | Low-resource deployment, opacity, single-centre validation | Pruning, distillation, quantisation and runtime acceleration; Grad-CAM++ and attention reporting; cross-institution generalisation testing | Embedded Systems; Software Engineering | Deployable prototype with a clinical simulation report |

Two features of this table deserve comment. First, the bottleneck column is not decorative. Each entry names a problem for which the field has candidate solutions but no consensus, and this is what allows Tier III and Tier IV laboratory work to be genuinely open-ended: a student who improves tail-class performance by a defensible margin has done something the reference implementation does not do. Second, the prerequisite column enforces a scheduling constraint. Because prerequisites are distributed across the second and third undergraduate years, the tiers cannot form a single course; they are delivered as laboratory strands embedded in the courses named, with Tier V offered as a capstone.

The vertical dependency is deliberate and is, we would argue, the design's principal claim. A student encounters modality laziness in Tier III only because a joint representation was constructed in Tier I and fused in Tier II; encountering it earlier would be meaningless, and never encountering it at all, which is the current situation, is the deficit we set out to remove. The same logic applies to out-of-distribution detection, which is unintelligible to a student who has not first watched a model behave well in-distribution.

## 4. Laboratory Modules and Competency Mapping

The four tiers and the engineering layer are realised as six laboratory modules containing 32 units and 64 contact hours in total. Table 2 gives the distribution. Units are classified by cognitive demand using the revised Bloom taxonomy [17]: foundation units target the apply level, design units the analyse and evaluate levels, and translation units the create level.

Table 2. Distribution of the 32 laboratory units across modules, with contact hours and cognitive level.

| Module | Tier | Units | Contact hours | Cognitive level |
|---|---|---|---|---|
| M1 Multimodal data engineering: ultrasound denoising, MRI segmentation, pathology text encoding, DICOM/NIfTI handling, de-identification | I | 6 | 12 | Apply |
| M2 Dynamic graph representation learning: graph construction, topology bootstrapping, low-rank and sparse regularisation | I | 6 | 12 | Apply / Analyse |
| M3 Cross-modal fusion and reasoning: graph attention, gated recurrent fusion, missing-modality completion, fusion ablation | II | 7 | 14 | Analyse / Evaluate |
| M4 Knowledge evolution: orthogonal projection, prompt-based transfer, gradient modulation, incremental protocols | III | 5 | 10 | Analyse / Evaluate |
| M5 Long-tailed and out-of-distribution robustness: contrastive pretraining, adversarial re-weighting, OOD detection, abstention policy | IV | 4 | 8 | Evaluate |
| M6 Deployment and clinical simulation: pruning, quantisation, latency profiling, interpretability reporting, multi-centre verification | V | 4 | 8 | Create |
| Total | – | 32 | 64 | – |

Every unit is documented in a common format (objective, theoretical background, clinical rationale, procedure, expected observations, failure modes, extension questions and ethical notes) and is accompanied by a reference implementation that students may read but must not submit. The clinical rationale field is the one we found hardest to write, and it carries the largest share of the unit's educational intent: it states, in clinical rather than engineering language, why the technical quantity being manipulated matters to a patient. For the myometrial-invasion task that anchors several units, for instance, the rationale explains why the boundary between superficial and deep invasion changes surgical staging, which in turn explains why a symmetric error metric is the wrong choice.

Modules are differentiated by programme. Students in intelligent medical engineering take M1–M5 as required work and M6 as a capstone, emphasising the algorithm-to-clinic pathway. Students in biomedical engineering take M1, M3 and M6 as required work, with additional emphasis on device adaptation and edge-terminal debugging, reflecting a different professional destination. Clinical and pharmacy students take a reduced M1 and a specially written subset of M3 and M6 focused on interrogating and interpreting model output rather than building it. We consider this distinction important, since the competency a future clinician needs is critical consumption rather than implementation.

## 5. Platform Implementation

A virtual clinical diagnosis workstation has four layers. The data layer uses a de-identified corpus from at least three institution types to support multi-centre exercises. The modelling layer encapsulates graph architectures and interpretability instrumentation so students write only the component under study. The deployment layer provides pruning, distillation, and 8-bit quantisation targeting ultrasound-cart-class processors, with a specification of 60 per cent compression and 80 ms latency. The reporting layer renders Grad-CAM++ heat maps and requires a clinical-language rationale document. Supporting infrastructure includes a simulation laboratory for fifty students, GPU servers, and a technical team ensuring 98 per cent availability. Teaching staff must pass internal certification.

## 6. Staged Delivery and Assessment

Delivery follows a three-stage progression (foundation, design and translation) that cuts across the module structure and defines what a student should be able to do at each point. Assessment is criterion-referenced throughout; thresholds are published in advance, and students are told that a well-executed experiment reporting a negative result is preferable to an unexamined positive one. Table 3 states the scheme.

Table 3. Three-stage delivery with representative deliverables, criterion-referenced thresholds and weighting.

| Stage | Representative deliverable | Criterion (design specification) | Weighting |
|---|---|---|---|
| Foundation (M1–M2) | Unified multimodal encoding of an ultrasound, MRI and pathology-text triple; reproducible preprocessing pipeline | Encoding correctness at or above 90 per cent on the held-out verification set; pipeline reproducible from the submitted configuration alone | 30 per cent |
| Design (M3–M5) | Team prototype of a multimodal diagnostic system, with a proposed algorithmic modification and its ablation | Modification yields at least a 10 per cent relative F1 gain on the designated tail class, or an equally rigorous documented negative result with a stated hypothesis for the failure | 40 per cent |
| Translation (M6) | Compressed model deployed to the edge target; multi-centre verification report; clinical simulation write-up | Latency no greater than 80 ms on the edge target at the specified compression ratio; cross-institution performance drop quantified with confidence intervals and explained | 30 per cent |

Three aspects of this scheme warrant justification. The first is the admission of documented negative results at the design stage. In a curriculum that transfers open research problems, requiring a positive result would force students either to select safe modifications or to report selectively, and both defeat the purpose; what is assessed instead is the soundness of the experiment and the reasoning about why it behaved as it did. The second is the insistence on confidence intervals in the translation stage, since a single point estimate of cross-institution performance is not interpretable. The third is the weight given to translation: assigning 30 per cent to work that produces no accuracy improvement whatsoever signals that deployability and validation are first-class engineering outcomes.

Evaluation of the system itself, as distinct from evaluation of students, will follow a mixed-methods design across the two implementation cycles: pre- and post-instruction concept inventories covering multimodal fusion and generalisation; blinded rubric scoring of design-stage and translation-stage artefacts; a validated self-efficacy instrument; and semi-structured interviews with students and with the clinical staff who review the simulation reports. Comparison will be against the preceding cohort taught under the legacy single-modality laboratory sequence, with the well-known limitations of a historical control acknowledged rather than concealed. We state the plan here so that the design can be judged against the evidence it proposes to generate.

## 7. Data Governance, Safety and Ethics

Data are tiered into public benchmarks, de-identified institutional data, and original records. Students access only the first two tiers under role-based access and confidentiality agreements. De-identification is taught as a lab unit, not performed invisibly. The virtual workstation ensures errors are recoverable and no patient is exposed. Operating rules, usage registration, and audit logs complete the governance model.

## 8. Conclusion

We have presented the design of an experimental teaching system that converts an active multimodal medical imaging research programme into an undergraduate laboratory curriculum of 32 units and 64 contact hours, organised as four progressive tiers plus an engineering layer, delivered on a virtual clinical diagnosis workstation and assessed against published criterion-referenced thresholds. It responds to three separable deficits in current practice (modality, authenticity and deployment) with three separable remedies: in content, in data and assessment, and in infrastructure. Its distinguishing commitment is that every tier is anchored to a bottleneck that remains open in the research literature, which makes genuinely open-ended laboratory work possible and gives the curriculum a structure that does not expire with the current generation of tooling. Empirical evaluation is under way and will be reported separately.

## Acknowledgment

This work is supported by the Innovative Experimental Project Programme for Higher Education Institutions of the Sichuan Provincial Department of Education. The authors thank the clinical staff of the Affiliated Hospital of Southwest Medical University for their guidance on the clinical rationale documentation.